\documentclass[10pt, conference]{IEEEtran}
\usepackage{caption,subfig}
\usepackage{textcomp}
\usepackage{epsfig,graphicx, graphics}
\usepackage{xcolor}
\usepackage{amsfonts,amsmath,amssymb}
\usepackage{fixltx2e} 
\usepackage{booktabs}
\makeatletter
\let\@currsize\normalsize
\makeatother
\usepackage{setspace}
\usepackage[ruled,vlined]{algorithm2e}
\usepackage{soul} 
\usepackage{tablefootnote}
\usepackage{multirow}
\usepackage{gensymb}
\usepackage{hyperref}
\usepackage{url}
\usepackage{wrapfig}

\begin{document}

\title{Multi-modal Sensor Registration for Vehicle Perception via Deep Neural Networks}
\author{Michael Giering, Vivek Venugopalan and Kishore Reddy\\
United Technologies Research Center\\
E. Hartford, CT 06018, USA \\
Email: \{gierinmj, venugov, reddykk\}@utrc.utc.com}

\maketitle
\begin{abstract}

The ability to simultaneously leverage multiple modes of sensor information is critical for perception of an automated vehicle's physical surroundings. Spatio-temporal alignment of registration of the incoming information is often a prerequisite to analyzing the fused data. The persistence and reliability of multi-modal registration is therefore the key to the stability of decision support systems ingesting the fused information. LiDAR-video systems like on those many driverless cars are a common example of where keeping the LiDAR and video channels registered to common physical features is important. We develop a deep learning method that takes multiple channels of heterogeneous data, to detect the misalignment of the LiDAR-video inputs.  A number of variations were tested on the Ford LiDAR-video driving test data set and will be discussed. To the best of our knowledge the use of multi-modal deep convolutional neural networks for dynamic real-time LiDAR-video registration has not been presented.

\end{abstract}

\section{Motivation} 
\label{sec:motivation}

Navigation and situational awareness of optionally manned vehicles requires the integration of multiple sensing modalities such as Light Detection and Ranging (LiDAR) and video, but could just as easily be extended to other modalities including Radio Detection And Ranging (RADAR), Short-Wavelength Infrared (SWIR) and Global Positioning System (GPS). Spatio-temporal registration of information from multi-modal sensors is technically challenging in its own right. For many tasks such as pedestrian and object detection tasks that make use of multiple sensors, decision support methods rest on the assumption of proper registration. Most approaches \cite{Bodensteiner2012Real-time-} in LiDAR-video for instance, build separate vision and LiDAR feature extraction methods and identify common anchor points in both. Alternatively, by generating a single feature set on LiDAR, Video and optical flow, it enables the system to to capture mutual information among modalities more efficiently. The ability to dynamically register information from the available data channels for perception related tasks can alleviate the need for anchor points \emph{between} sensor modalities. We see auto-registration as a prerequisite need for operating on multi-modal information with confidence.

Deep neural networks (DNN) lend themselves in a seamless manner for data fusion on time series data. For some challenges in which the modalities share significant mutual information, the features generated on the fused information can provide insight that neither input alone can \cite{Ngiam2011Multimodal}. In effect the ML version of, "the whole is greater than the sum of it's parts". 

Autonomous navigation places significant constraints on the speed of perception algorithms and their ability to drive decision making in real-time. Though computationally intensive to train, our implemented DCNN run easily within our real-time frame rates of 8 fps and could accommodate more standard rates of 30 fps. 
With most research in deep neural networks focused on algorithmic improvements and novel applications, a significant benefit to applied researchers is sometimes under appreciated. The automated feature generation of DNNs enables us to create mutli-modal systems with far less overhead. The need for domain experts and hand-crafted feature design are lessened, allowing more rapid prototyping and testing. The generalization of auto-registration across multiple assets is clearly a path to be explored. 

In this paper, the main contributions are: (i) formulation of an image registration problem as a fusion of modalities from different sensors, namely LIDAR (L), video (Grayscale or R,G,B) and optical flow (U,V); (ii) performance evaluation of deep convolutional neural networks (DCNN) with various input parameters, such as kernel filter size and different combinations of input channels (R,G,B,Gr,L,U,V); (iii) fusion of patch-level and image-level predictions to generate alignment at the frame-level. The experiments were conducted using a publicly available dataset from FORD and the University of Michigan \cite{Pandey2011Ford-Campu}. The DCNN implementation was executed on an NVIDIA Tesla K40 GPU with 2880 cores and compute power of 5 TFLOPS (single precision). The paper is organized into the following sections: Section \ref{sec:motivation} describes the introduction and motivation for this work; Section \ref{sec:previous_work} provides a survey of the related work; the problem formulation along with the dataset description and the preprocessing is explained in Section \ref{sec:problem_statement}; Section \ref{sec:model_description} gives the details of the DCNN setup for the different experiments; Section \ref{sec:experiments} describes the experiments and the post-processing steps for visualizing the qualitative results; finally Section \ref{sec:conclusions_and_future_work} summarizes the paper and concludes with future research thrusts.


\section{Previous Work} 
\label{sec:previous_work}

A great amount has been published on various multi-modal fusion methods \cite{Ross2003Informatio}, \cite{Gregor2011Learning-R}, \cite{Wu2004Optimal-Mu}, \cite{Snoek2006The-Challe}. The most common approaches taken generate features of interest in each modality separately and create a decision support mechanism that aggregates features across modalities. If spatial alignment is required across modalities, as it is for LiDAR-video such filter methods \cite{Thrun2011Googles-dr} are required to ensure proper inter-modal registration. These filter methods for leveraging 3D LiDAR and 2D images are often geometric in nature and make use of projections between the different data spaces. 

Automatic registration of 2D video and 3D LiDAR has been a widely researched topic for over a decade \cite{ Wang2009A-Robust-A}, ~\cite{Kim2014Automatic-}, ~\cite{Mastin2009Automatic-}, ~\cite{Bodensteiner2012Real-time-}. Its application in real-time autonomous navigation makes it a challenging problem. Majority of the 2D-3D registration algorithms are based on feature matching. Geometric features like corners and edges are extracted from detected vanishing points ~\cite{Liu2007-Vanishing-points},~\cite{Ding2008-Vanishing-point}, line segments ~\cite{Frueh2004-Linesegment}, ~\cite{Stamos2008-Linesegment}, and shadows ~\cite{Troccoli2004-ashadow}. Feature based approaches generally rely on dense 3D point cloud and additional knowledge of relative sun position and GPS/inertial navigation system (INS). Another approach used for video and LiDAR auto-registration is to reconstruct 3D point cloud from video sequences using structure from motion (SFM) and performing 3D-3D registration ~\cite{Zhao2004-alignment-3Dcloud}, ~\cite{Liu2006-alignment-sfm}. 3D-3D registration is more difficult and computationally expensive compared to 2D-3D registration. 

The use of deep neural networks to analyze multi-modal sensor inputs  has increased sharply in just the last few years, including audio-video \cite{Ngiam2011Multimodal}, \cite{Kim2013Deep-Learn}, image/text \cite{Srivastava2012Multimodal}, image/depth \cite{Lenz2013Deep-Learn} and LiDAR-video To the best of our knowledge the use of multi-modal deep neural networks for dynamic LiDAR-video registration has not been presented.

A common challenge for data fusion methods is deciding at what level features from the differing sensor streams should be brought together. The deep neural network (DNN) approach most similar to the more traditional data fusion methods is to train DNNs independently on sensor modalities and then use the high-level outputs of those networks as inputs to a subsequent aggregator, which could also be a DNN. This is analogous to the earlier example of learning 3D/2D features and the process of identifying common geometric features. 

It is possible however to apply DNNs with a more agnostic view enabling a unified set of features to be learned across multi-modal data. In these cases the input channels aren't differentiated. Unsupervised methods including deep Boltzmann machines and deep auto-encoders for learning such joint representations have been successful.  

Deep convolutional neural networks (DCNN) enable a similar agnostic approach to input channels. A significant difference is that target data is required to train them as classifiers. This is the approach chosen by us for automating the registration of LiDAR-video and optical-flow, in which we are combining 1D/3D/2D data representations respectively to learn a unified model across as many as 6D.


\section{Problem Statement} 
\label{sec:problem_statement}

Being able to detect and correct the misalignment (registration, calibration) among sensors of the same or different kinds, is critical for decision support systems operating on their fused information streams. For our work DCNNs were implemented for the detection of small spatial misalignments in LiDAR and Video frames. The methodology is directly applicable to temporal registration as well. LiDAR-video data collected from a driverless car was chosen for the multi-modal fusion test case. LiDAR-video is a common combination for providing perception capabilities to many types of ground and airborne platforms including driverless cars \cite{Thrun2011Googles-dr}. 

\subsection{Ford LiDAR-video Dataset and Experimental Setup} 
\label{sub:ford_lidar_video_dataset_and_experimental_setup}


The FORD LiDAR-video dataset \cite{Pandey2011Ford-Campu} is collected by an autonomous Ford F-250 vehicle integrated with the following perception and navigation sensors as follows: 
\begin{itemize}
    \item Velodyne HDL-64E LiDAR with two blocks of lasers spinning at 10 Hz and a maximum range of 120m.
    \item Point Grey Ladybug3 omni-directional camera system with six 2-Mega-pixel cameras collecting video data at 8fps with $1600\times1600$ resolution.
    \item Two Riegl LMS-Q120 LIDAR sensors installed in the front of the vehicle generating range and intensity data when the laser sweeps its $80\degree$ field of view (FOV).
    \item Applanix POS-LV420 INS with Trimble GPS system providing the 6 degrees of freedom (DOF) estimates at 100 Hz.
    \item Xsens MTi-G sensor consisting of accelerometer, gyroscope, magnetometer, integrated GPS receiver, static pressure sensor and temperature sensor. It measures the GPS co-ordinates of the vehicle and also provides the 3D velocity and 3D rate of turn.
\end{itemize}

\begin{figure}[htbp]
    \centering
    \includegraphics[scale=0.35]{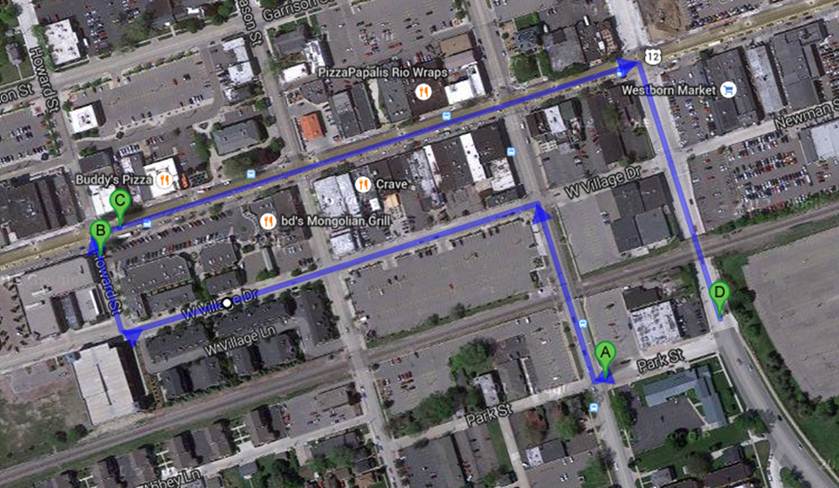}
    \caption{Training (A to B) and testing (C to D) tracks in the downtown Dearborn Michigan.}
    \label{fig:ford_train_test_track}
\end{figure}

This dataset is generated by the vehicle while driving in and around the Ford research campus and downtown Michigan. The data includes feature rich downtown areas as well as featureless empty parking lots. As shown in Figure \ref{fig:ford_train_test_track}, we divided the data set into training and testing sections A to B and C to D respectively. They were chosen in a manner that minimizes the likelihood of contamination between training and testing. Because of this, the direction of the light source is never the same in the testing and training sets.   


\subsection{Optical Flow} 
\label{sub:optical_flow}

\begin{figure}[htbp]
    \centering
    \includegraphics[scale=0.5]{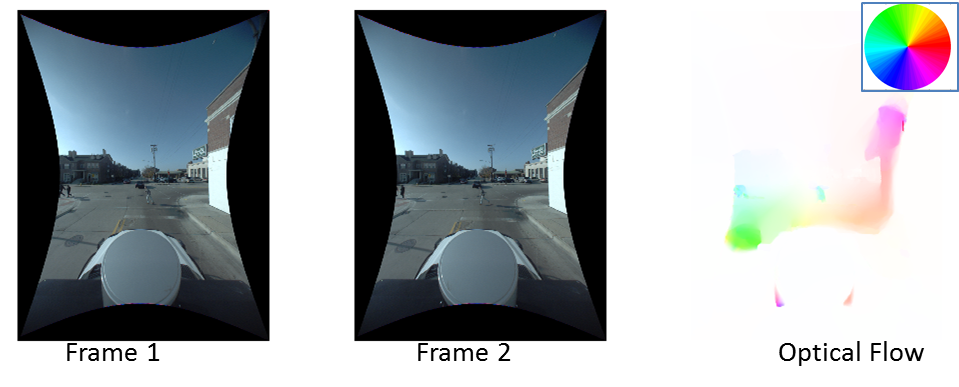}
    \caption{Optical flow: Hue indicates orientation and saturation indicates magnitude}
    \label{fig:Figures_OptFlow placeholder}
\end{figure}

In the area of navigation of mobile robots, optical flow has been widely used to estimate egomotion \cite{Prazdny1980-egomotion-OF}, depth maps \cite{Shahraray1988-depthestimation-OF}, reconstruct dynamic 3D scene depth \cite{Yang2012-reconstruction-OF}, and segment moving objects \cite{Shao2002-seg-OF}. Optical flow provides information of the scene dynamics and is expressed as an estimate of velocity at each pixel from two consecutive frames, denoted by $\vec{u}$ and $\vec{v}$. The motion field from these two frames is measured by the motion of the pixel brightness pattern, where the changes in image brightness is due to the camera or object motion. \cite{Liu2009Beyond-Pix} describes an algorithm for computing optical flow from images, which is used during the preprocessing step. Figure \ref{fig:Figures_OptFlow placeholder} shows an example of the optical flow computed using two consecutive frames from the Ford LiDAR-video dataset. By including optical flow as input channels, we imbue the DCNN with information on the dynamics observed across time steps.


\subsection{Preprocessing} 
\label{sub:preprocessing}
\begin{figure}[htbp]
    \centering
    \subfloat[Visualization of the elliptically distributed $N=9$ classes]
    {
        \includegraphics[scale=0.4]{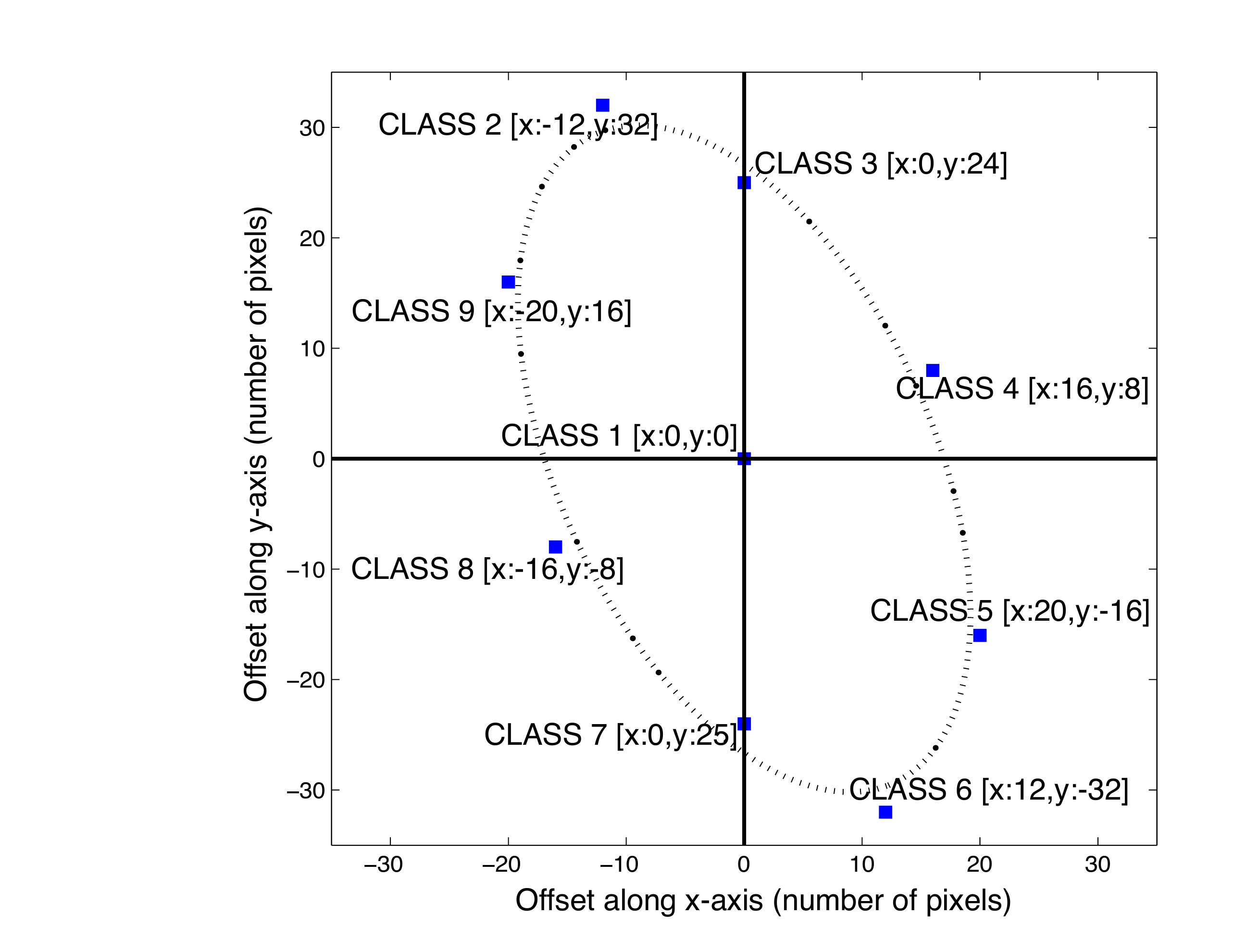}
        \label{fig:Figures_Ellipse}
    }
    \hspace{0.2cm}
    \subfloat[Stacking of $p\times p$ patch over all the different channels]
    {
        \includegraphics[scale=0.5]{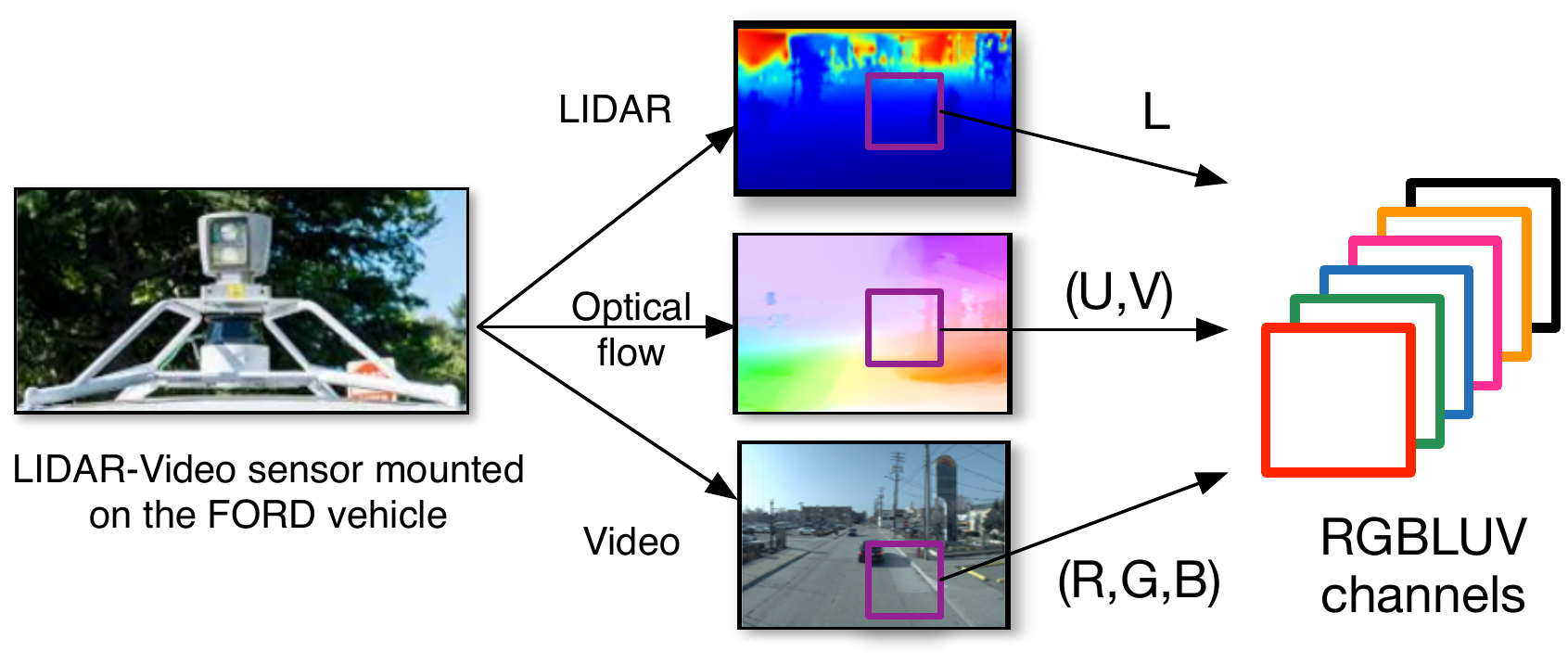}
        \label{fig:ImageChStride}
    }
    \caption{Preprocessing steps}
    \label{fig:preprocessing_steps}
\end{figure}

At each video frame timestep, the inputs to our model consist of \emph{C} channels of data with \emph{C} ranging from 3-6 channels. Channels consist of grayscale \emph{Gr} or \emph{(R,G,B)} information from the video, horizontal and vertical components of optical flow \emph{(U,V)} and depth information \emph{L} from LiDAR The data from each modality is reshaped to a fixed size of $800\times256$ values, which are partitioned into $p\times p$ patches at a prescribed stride. Each patch $p\times p$ is stacked across \emph{C} channels, effectively generating a vector of \emph{C} dimensions. The different preprocessing parameters are denoted by patch size \emph{p}, stride \emph{s} and the number of input channels \emph{C}.

Preprocessing is repeated \emph{N} times, where \emph{N} is the number of offset classes. For each offset class, the video (R,G,B) and optical flow (U,V) channels are kept static and the depth (L) channel from the LiDAR is moved by the offset simulating a misalignment between the video and the LiDAR sensors. In order to accurately detect the misalignment in the LiDAR and Video sensor data, a threshold is set to limit the information available in each channel. The LiDAR data has regions of sparsity and hence the LiDAR patches with a variance (${\sigma}^2 < 15\%$) are dropped from the final dataset. This leads to the elimination of the majority of foreground patches in the data set, reducing the size of the training and testing set by approximately $80\%$. Figure \ref{fig:Figures_Ellipse} shows a $N = 9$ class elliptically distributed set of offsets and Figure \ref{fig:ImageChStride} shows a $p\times p$ patch stacked across all the different \emph{C} channels.



\section{Model Description} 
\label{sec:model_description}

\begin{figure*}[htbp]
    \centering
    \includegraphics[scale=0.5]{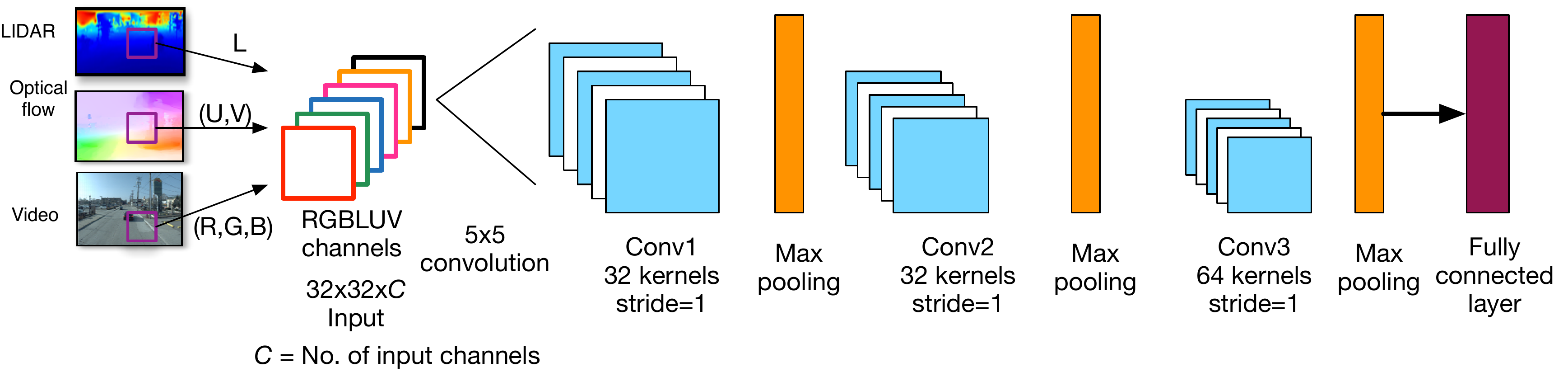}
    \caption{Experimental setup of the LiDAR-video DCNN with $5\times5$ convolution}
    \label{fig:Figures_lidar_dcnn_setup1}
\end{figure*}

Our models for auto-registration are DCNNs trained to classify the current misalignment of the LiDAR-video data streams into one of a predefined set of offsets. DCNNs are probably the most successful deep learning model to date on fielded applications. The fact that the algorithm shares weights in the training phase, results in fewer model parameters and more efficient training. DCNNs are particularly useful for problems in which local structure is important, such as object recognition in images and temporal information for voice recognition. The alternating steps of convolution and pooling generates features at multiple scales which in turn imbues DCNN's with scale invariant characteristics.

The model shown in Figure \ref{fig:Figures_lidar_dcnn_setup1} consists of 3 pairs of convolution-pooling layers, that estimates the offset between the LiDAR-video inputs at each time step. For each patch within a timestep, there are $N$ variants with the LiDAR-video-optical flow inputs offset by the predetermined amounts. The CNN outputs to a softmax layer, thereby providing an offset classification value for each patch of the frame. As described in Section \ref{sub:preprocessing}, $32\times32$ patches were stacked across the different channels and provided as the input to the DCNN. All the $6$ channels \emph{RGBLUV} were used for the majority of the experiments, whereas only $4$ channels were required for the \emph{RGBL} and the \emph{GrLUV} experiments. The first convolutional layer uses $32$ filters (or kernels) of size $5 \times 5 \times \mathit{C} $ with a stride of $1$ pixel and padding of $2$ pixels on the edges. The following pooling layer generates the input data (of size $16 \times 16 \times 32$) for the second convolutional layer. This layer uses $32$ filters of size $5 \times 5 \times 32$ with a stride of $1$ pixel and padding of $2$ pixels on the edges. A second pooling layer, similar to the first one is used to generate input with size $8 \times 8 \times 32$ for the third convolutional layer that uses $64$ filters of size $5 \times 5 \times 32$ with the stride and padding same as previous convolutional layer. The third pooling layer with similar configuration as the two previous pooling layers connects to an output softmax layer with labels corresponding to the $N=9$ classes. The DCNN described above was trained using stochastic gradient descent with a mini-batch size of $100$ epochs. The DCNN is configured with Rectified Linear Units (ReLUs), as they train several times faster than their equivalents with $\tanh$ connections \cite{Nair2010Rectified-}

The NVIDIA Kepler series K40 GPUs \cite{NVIDIA-Inc.2012NVIDIAs-Ne} are very FLOPS/Watt efficient and are being used to drive real-time image processing capabilities \cite{Venugopal2013Accelerati}. These GPUs consist of 2880 cores with 12 GB of on-board device memory (RAM). Deep Learning applications have been targeted on GPUs previously in \cite{Krizhevsky2012Imagenet-C} and these implementations are both compute and memory bound. Stacking of channels results in a vector of $32 \times 32 \times \mathit{C}$, which is suitable for the Single Instruction Multiple Datapath (SIMD) architecture of the GPUs. At the same time, the training batch size caches in the GPU memory, so the utilization of the K40 GPU's memory is very high. This also results in our experiments to run successfully on a single GPU instead of partitioning the different layers over multiple GPUs.
 


\section{Experiments} 
\label{sec:experiments}

\subsection{Dataset using elliptically distributed offsets} 
\label{sub:dataset_using_elliptically_distributed_offsets}

\begin{figure*}[htbp]
    \centering
    \includegraphics[scale=0.35]{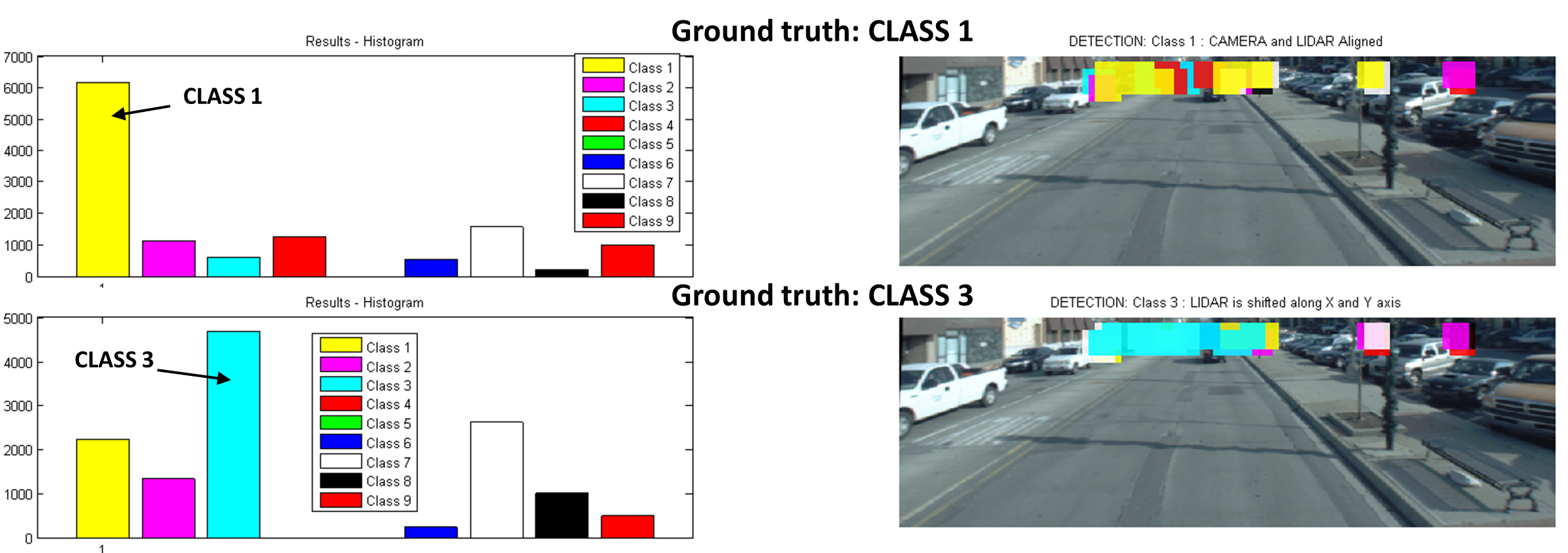}
    \caption{Left: Histogram of the classes detected by all patches. Right: Location of patches color coded with the predicted class}
    \label{fig:Figures_Voting}
\end{figure*}

In our experiments, elliptically distributed set of $N = 9$ offsets of the LiDAR-video data were considered. The LiDAR data is displaced along an ellipse with a major axis of $32$ pixels and a minor axis of $16$ pixels rotated clockwise from x-axis by $45\degree$ as shown in Figure \ref{fig:Figures_Ellipse}. Separate training and testing sets were generated from two different tracks as shown in Figure \ref{fig:ford_train_test_track} for all the $N = 9$ offsets of LiDAR data. Training and testing tracks have never seen regions and also have different lighting conditions. Our preprocessing step described in Section~\ref{sub:preprocessing} results in $223,371$ and $126,513$ patches for testing and training extracted from $469$ and $224$ images respectively.

In the testing phase, for each frame a simple voting scheme is used to aggregate the patch level offset predictions to a single frame level prediction. A sample histogram of the patch level predictions is show in Figure~\ref{fig:Figures_Voting}. We color each patch of the frame with a color corresponding to the predicted class as shown in Figure~\ref{fig:Figures_Voting}.


\subsection{Experimental results} 
\label{sub:experimental_results}
\begin{figure}[htbp]
    \centering
    \subfloat[Patch level confusion matrix (41.05\% accuracy)]
    {
        \includegraphics[scale=0.42]{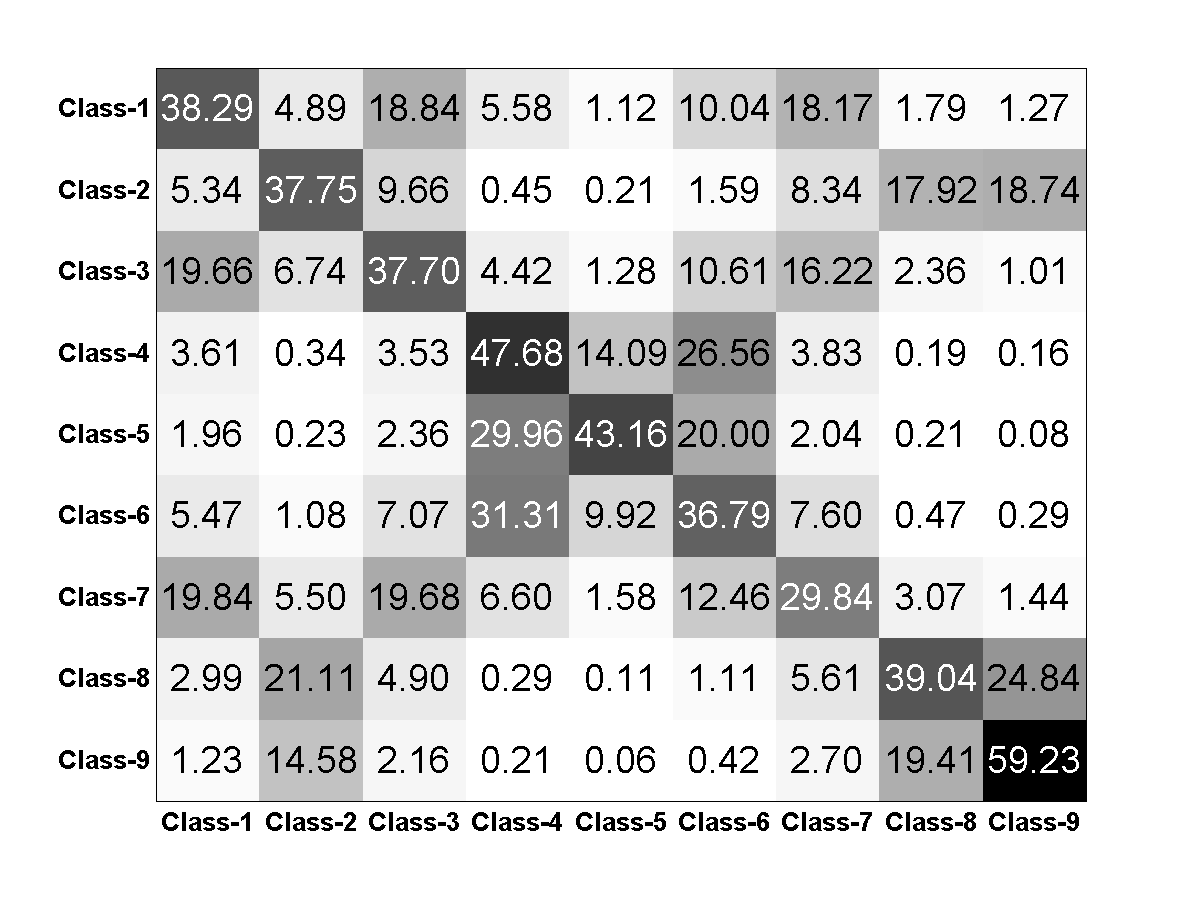}
        \label{fig:patch_level_cm_GrLUV}
    }
    \hspace{0.2cm}
    \subfloat[Image level confusion matrix (76.69\% accuracy)]
    {
        \includegraphics[scale=0.42]{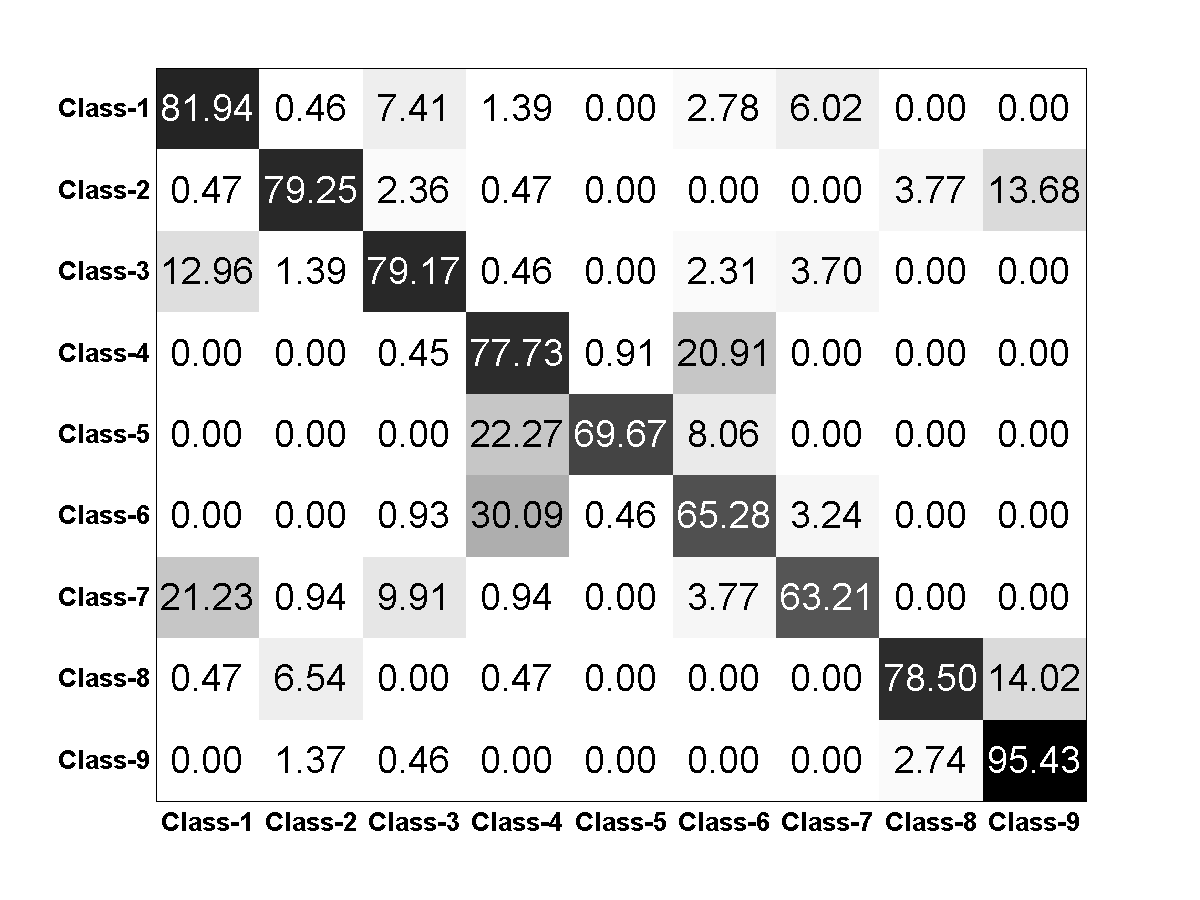}
        \label{fig:image_level_cm_GrLUV}
    }
    \caption{Confusion Matrix for elliptically distributed $N=9$ classes using Greyscale, Optical Flow and LiDAR channels with a filter size of 9}
    \label{fig:confusionmatrix}
\end{figure}

Table~\ref{table:cnn_param} lists the inputs and CNN parameters explored ranked in the order of increasing accuracy. We averaged the values across the diagonal of the confusion matrix to determine the image level and patch level accuracy. Patch level accuracy is the individual performance of all the $32\times32$ patches from the testing images. Classification of patches belonging to a single time-step are voted to predict the shift for image level accuracy. In Table~\ref{table:cnn_param}, the first 3 columns show the results for different number of filter combinations in the convolutional layers with fixed number of filters and input channels \emph{RGBLUV}. We observed that the image  and patch level accuracy decreased with the increase in the number of filters. For experiments shown in columns 4 and 5, the filter size was increased, with the number of filters constant at $(32,32,64)$. We observed that for the 6 channels \emph{RGBLUV}, filter size of 9 gave the best image level accuracy of $63.03\%$. Column 6 shows the results of our experiment after dropping the optical flow \emph{UV} channels. The image and patch level accuracy decreased for this case, indicating that optical flow contributed significantly towards image registration. The remaining experiments utilized the Grayscale information \emph{Gr} instead of \emph{RGB} and produced the best results with $76.69\%$ and $41.05\%$ image and patch level accuracy respectively. Table~\ref{table:temporal_performance} shows that by using information from consecutive frames the performance increases significantly.

\begin{table*}[htbp]
\small
\caption{Results of different combination of input channels \emph{C} and CNN parameters}

\begin{center}
\begin{tabular}{|p{2cm}|c|c|c|c|c|c|c|c|} \hline
    \textbf{Channels} & \multicolumn{5}{|c|}{RGBLUV} & RGBL & \multicolumn{2}{|c|}{GrLUV} \\
    \hline
    \textbf{Filter size} &\multicolumn{3}{|c|}{5} & 7 & 9 & \multicolumn{2}{|c|}{5} & 9 \\
    \hline
    \textbf{\# of filters} & (32,32,32) & (32,32,64) & (64,64,64) & \multicolumn{5}{|c|}{(32,32,64)}  \\
    \hline
    \textbf{Image Level accuracy(\%)} & 61.75 & 61.06 & 60.09 & 61.79 & 63.03 & 60.66 & 68.03 & \textbf{76.69} \\
    \hline
    \textbf{Patch Level accuracy(\%)} & 38.74 & 38.57 & 38.49 & 38.03 & 39.00 & 39.28 & 40.96 & \textbf{41.05} \\
    \hline
\end{tabular}
\label{table:cnn_param}
\end{center}
\end{table*}

\begin{table*}[htbp]
\small
\caption{Performance using temporal information}
\begin{center}
     \begin{tabular}{ | p{4cm} | l | l | l | l | l | l | l | l |} \hline
     \textbf{Number of consecutive time-steps used} & \textbf{1} & \textbf{2} & \textbf{3} & \textbf{4} & \textbf{5} & \textbf{6} & \textbf{7} & \textbf{8} \\ \hline
     \textbf{Accuracy(\%)} & 76.33 & 85.42 & 88.88 & 90.30 & 92.52 & 93.85 & 94.29 & 95.12\\ \hline
     \end{tabular}
     \label{table:temporal_performance}
\end{center}
\end{table*}



\section{Conclusions and Future Work} 
\label{sec:conclusions_and_future_work}
In this paper, we proposed a deep learning method to do LiDAR-Video registration. We demonstrated the effect of filter size, number of filters and different channels. We also showed the advantage of using temporal information, optical flow and grayscale.
The next step in taking this work forward is to complete our development of a deep auto-registration method for ground and aerial platforms requiring no a priori calibration ground truth.  Our aerospace applications in particular present noisier data with an increased number of degrees of freedom. The extension of these methods to simultaneously register information across multiple platforms and larger numbers of modalities will provide interesting challenges that we look forward to working on. 



\bibliographystyle{IEEEtran}
\bibliography{references}

\begin{thebibliography}{10}
\providecommand{\url}[1]{#1}
\csname url@rmstyle\endcsname
\providecommand{\newblock}{\relax}
\providecommand{\bibinfo}[2]{#2}
\providecommand\BIBentrySTDinterwordspacing{\spaceskip=0pt\relax}
\providecommand\BIBentryALTinterwordstretchfactor{4}
\providecommand\BIBentryALTinterwordspacing{\spaceskip=\fontdimen2\font plus
\BIBentryALTinterwordstretchfactor\fontdimen3\font minus
  \fontdimen4\font\relax}
\providecommand\BIBforeignlanguage[2]{{%
\expandafter\ifx\csname l@#1\endcsname\relax
\typeout{** WARNING: IEEEtran.bst: No hyphenation pattern has been}%
\typeout{** loaded for the language `#1'. Using the pattern for}%
\typeout{** the default language instead.}%
\else
\language=\csname l@#1\endcsname
\fi
#2}}

\bibitem{Bodensteiner2012Real-time-}
C.~Bodensteiner and M.~Arens, ``{Real-time 2D Video 3D LiDAR Registration},''
  in \emph{Pattern Recognition (ICPR), 2012 21st International Conference
  on}.\hskip 1em plus 0.5em minus 0.4em\relax IEEE, 2012, pp. 2206--2209.

\bibitem{Ngiam2011Multimodal}
J.~Ngiam, A.~Khosla, M.~Kim, J.~Nam, H.~Lee, and A.~Y. Ng, ``{Multimodal Deep
  Learning},'' in \emph{Proceedings of the 28th International Conference on
  Machine Learning (ICML-11)}, 2011, pp. 689--696.

\bibitem{Pandey2011Ford-Campu}
G.~Pandey, J.~R. McBride, and R.~M. Eustice, ``{Ford Campus Vision And Lidar
  Data Set},'' \emph{The International Journal of Robotics Research}, vol.~30,
  no.~13, pp. 1543--1552, 2011.

\bibitem{Ross2003Informatio}
A.~Ross and A.~Jain, ``{Information Fusion In Biometrics},'' \emph{Pattern
  recognition letters}, vol.~24, no.~13, pp. 2115--2125, 2003.

\bibitem{Gregor2011Learning-R}
K.~Gregor and Y.~LeCun, ``{Learning Representations By Maximizing
  Compression},'' \emph{arXiv preprint arXiv:1108.1169}, 2011.

\bibitem{Wu2004Optimal-Mu}
Y.~Wu, E.~Y. Chang, K.~C.-C. Chang, and J.~R. Smith, ``{Optimal Multimodal
  Fusion For Multimedia Data Analysis},'' in \emph{Proceedings of the 12th
  annual ACM international conference on Multimedia}.\hskip 1em plus 0.5em
  minus 0.4em\relax ACM, 2004, pp. 572--579.

\bibitem{Snoek2006The-Challe}
C.~G. Snoek, M.~Worring, J.~C. Van~Gemert, J.-M. Geusebroek, and A.~W.
  Smeulders, ``{The Challenge Problem For Automated Detection Of 101 Semantic
  Concepts In Multimedia},'' in \emph{Proceedings of the 14th annual ACM
  international conference on Multimedia}.\hskip 1em plus 0.5em minus
  0.4em\relax ACM, 2006, pp. 421--430.

\bibitem{Thrun2011Googles-dr}
S.~Thrun, ``{Google's driverless car},'' \emph{Ted Talk, Ed}, 2011.

\bibitem{Wang2009A-Robust-A}
L.~Wang and U.~Neumann, ``{A Robust Approach For Automatic Registration Of
  Aerial Images With Untextured Aerial LIDAR Data},'' in \emph{Computer Vision
  and Pattern Recognition, 2009. CVPR 2009. IEEE Conference on}.\hskip 1em plus
  0.5em minus 0.4em\relax IEEE, 2009, pp. 2623--2630.

\bibitem{Kim2014Automatic-}
H.~Kim, C.~D. Correa, and N.~Max, ``Automatic registration of lidar and optical
  imagery using depth map stereo,'' in \emph{Computational Photography (ICCP),
  2014 IEEE International Conference on}.\hskip 1em plus 0.5em minus
  0.4em\relax IEEE, 2014, pp. 1--8.

\bibitem{Mastin2009Automatic-}
A.~Mastin, J.~Kepner, and J.~Fisher, ``{Automatic Registration of LIDAR and
  Optical Images of Urban Scenes},'' in \emph{Computer Vision and Pattern
  Recognition, 2009. CVPR 2009. IEEE Conference on}.\hskip 1em plus 0.5em minus
  0.4em\relax IEEE, 2009, pp. 2639--2646.

\bibitem{Liu2007-Vanishing-points}
L.~Liu and I.~Stamos, ``A systematic approach for 2d-image to 3d-range
  registration in urban environments,'' in \emph{Computer Vision, 2007. ICCV
  2007. IEEE 11th International Conference on}, Oct 2007, pp. 1--8.

\bibitem{Ding2008-Vanishing-point}
M.~Ding, K.~Lyngbaek, and A.~Zakhor, ``Automatic registration of aerial imagery
  with untextured 3d lidar models,'' in \emph{Computer Vision and Pattern
  Recognition, 2008. CVPR 2008. IEEE Conference on}, June 2008, pp. 1--8.

\bibitem{Frueh2004-Linesegment}
C.~Frueh, R.~Sammon, and A.~Zakhor, ``Automated texture mapping of 3d city
  models with oblique aerial imagery,'' in \emph{3D Data Processing,
  Visualization and Transmission, 2004. 3DPVT 2004. Proceedings. 2nd
  International Symposium on}, Sept 2004, pp. 396--403.

\bibitem{Stamos2008-Linesegment}
\BIBentryALTinterwordspacing
I.~Stamos, L.~Liu, C.~Chen, G.~Wolberg, G.~Yu, and S.~Zokai,
  ``\BIBforeignlanguage{English}{Integrating automated range registration with
  multiview geometry for the photorealistic modeling of large-scale scenes},''
  \emph{\BIBforeignlanguage{English}{International Journal of Computer
  Vision}}, vol.~78, no. 2-3, pp. 237--260, 2008. [Online]. Available:
  \url{http://dx.doi.org/10.1007/s11263-007-0089-1}
\BIBentrySTDinterwordspacing

\bibitem{Troccoli2004-ashadow}
A.~J. Troccoli and P.~K. Allen, ``A shadow based method for image to model
  registration,'' in \emph{In IEEE Workshop on Image and Video Registration,
  Conf. on Comp. Vision and}, 2004.

\bibitem{Zhao2004-alignment-3Dcloud}
W.~Zhao, D.~Nister, and S.~Hsu, ``Alignment of continuous video onto 3d point
  clouds,'' in \emph{Computer Vision and Pattern Recognition, 2004. CVPR 2004.
  Proceedings of the 2004 IEEE Computer Society Conference on}, vol.~2, June
  2004, pp. II--II.

\bibitem{Liu2006-alignment-sfm}
L.~Liu, I.~Stamos, G.~Yu, G.~Wolberg, and S.~Zokai, ``Multiview geometry for
  texture mapping 2d images onto 3d range data,'' in \emph{Computer Vision and
  Pattern Recognition, 2006 IEEE Computer Society Conference on}, vol.~2, June
  2006, pp. 2293--2300.

\bibitem{Kim2013Deep-Learn}
Y.~Kim, H.~Lee, and E.~M. Provost, ``{Deep Learning for Robust Feature
  Generation in Audiovisual Emotion Recognition},'' in \emph{Acoustics, Speech
  and Signal Processing (ICASSP), 2013 IEEE International Conference on}.\hskip
  1em plus 0.5em minus 0.4em\relax IEEE, 2013, pp. 3687--3691.

\bibitem{Srivastava2012Multimodal}
N.~Srivastava and R.~Salakhutdinov, ``{Multimodal Learning With Deep Boltzmann
  Machines},'' in \emph{Advances in neural information processing systems},
  2012, pp. 2222--2230.

\bibitem{Lenz2013Deep-Learn}
I.~Lenz, H.~Lee, and A.~Saxena, ``{Deep Learning for Detecting Robotic
  Grasps},'' \emph{arXiv preprint arXiv:1301.3592}, 2013.

\bibitem{Prazdny1980-egomotion-OF}
\BIBentryALTinterwordspacing
K.~Prazdny, ``\BIBforeignlanguage{English}{Egomotion and relative depth map
  from optical flow},'' \emph{\BIBforeignlanguage{English}{Biological
  Cybernetics}}, vol.~36, no.~2, pp. 87--102, 1980. [Online]. Available:
  \url{http://dx.doi.org/10.1007/BF00361077}
\BIBentrySTDinterwordspacing

\bibitem{Shahraray1988-depthestimation-OF}
B.~Shahraray and M.~Brown, ``Robust depth estimation from optical flow,'' in
  \emph{Computer Vision., Second International Conference on}, Dec 1988, pp.
  641--650.

\bibitem{Yang2012-reconstruction-OF}
Y.~{Yang}, Q.~{Liu}, R.~{Ji}, and Y.~{Gao}, ``{Dynamic 3D Scene Depth
  Reconstruction via Optical Flow Field Rectification},'' \emph{PLoS ONE},
  vol.~7, p. 47041, Nov. 2012.

\bibitem{Shao2002-seg-OF}
S.-Y. Chien, S.-Y. Ma, and L.-G. Chen, ``Efficient moving object segmentation
  algorithm using background registration technique,'' \emph{Circuits and
  Systems for Video Technology, IEEE Transactions on}, vol.~12, no.~7, pp.
  577--586, Jul 2002.

\bibitem{Liu2009Beyond-Pix}
C.~Liu, ``{Beyond Pixels: Exploring New Representations and Applications for
  Motion Analysis},'' Ph.D. dissertation, Massachusetts Institute of
  Technology, Cambridge, MA, USA, 2009, aAI0822221.

\bibitem{Nair2010Rectified-}
V.~Nair and G.~E. Hinton, ``{Rectified Linear Units Improve Restricted
  Boltzmann Machines},'' in \emph{Proceedings of the 27th International
  Conference on Machine Learning (ICML-10)}, 2010, pp. 807--814.

\bibitem{NVIDIA-Inc.2012NVIDIAs-Ne}
{NVIDIA Inc.}, ``{NVIDIA's Next Generation CUDA Compute Architecture: Kepler TM
  GK110},'' Whitepaper, May 2012.

\bibitem{Venugopal2013Accelerati}
V.~Venugopal and S.~Kannan, ``Accelerating real-time lidar data processing
  using gpus,'' in \emph{Circuits and Systems (MWSCAS), 2013 IEEE 56th
  International Midwest Symposium on}, August 2013, pp. 1168--1171.

\bibitem{Krizhevsky2012Imagenet-C}
A.~Krizhevsky, I.~Sutskever, and G.~E. Hinton, ``{Imagenet Classification With
  Deep Convolutional Neural Networks},'' in \emph{Advances in neural
  information processing systems}, 2012, pp. 1097--1105.

\end{thebibliography}

\end{document}